\documentclass[sigconf,authorversion]{acmart}

\usepackage{graphicx}

\AtBeginDocument{%
  }

\usepackage{framed}
\usepackage{csquotes}

\copyrightyear{2024}
\acmYear{2024}

\setcopyright{cc}
\setcctype[4.0]{by-sa}

\acmConference[EuroPLoP 2024]{29th European Conference on Pattern Languages of Programs, People, and Practices}{July 3--7, 2024}{Irsee, Germany} 
\acmBooktitle{29th European Conference on Pattern Languages of Programs, People, and Practices (EuroPLoP 2024), July 3--7, 2024, Irsee, Germany}
\acmDOI{10.1145/3698322.3698338} 
\acmISBN{979-8-4007-1683-6/24/07}

\newcommand{\stars}{\begin{center}$\ast\ast\ast$\end{center}}

\definecolor{shadecolor}{rgb}{0.8,0.8,0.8}

\begin{document}

\title{An Exploration of Pattern Mining with ChatGPT}

\author{Michael Weiss}
\email{michael_weiss@carleton.ca}
\orcid{0000-0001-7633-5286}
\affiliation{%
  \institution{Carleton University}
  \city{Ottawa}
  \country{Canada}
}

\begin{abstract}
This paper takes an exploratory approach to examine the use of ChatGPT for pattern mining. It proposes an eight-step collaborative process that combines human insight with AI capabilities to extract patterns from known uses. The paper offers a practical demonstration of this process by creating a pattern language for integrating Large Language Models (LLMs) with data sources and tools. LLMs, such as ChatGPT, are a new class of AI models that have been trained on large amounts of text, and can create new content, including text, images, or video. The paper also argues for adding affordances of the underlying components as a new element of pattern descriptions. The primary audience of the paper includes pattern writers interested in pattern mining using LLMs.
\end{abstract}




\maketitle

\section{Introduction}
The problem examined in this paper is whether ChatGPT (GPT 4) can be used to mine patterns.\footnote{The experiment was conducted in February 2024.} The paper is exploratory in nature. I am interested in learning how the pattern mining process can benefit from ChatGPT, and whether the results of this process are useful in practice. Large Language Models, (LLMs) such as ChatGPT, are a new class of AI models that have been trained on large amounts of text, and can generate new content, including text, images, or video \cite{bommasani2022opportunities}.

The paper makes three contributions. First, it shows how ChatGPT can be used to mine patterns from known uses. Second, it demonstrates the practical application of the process. Third, it argues for the addition of affordances as a new element of pattern descriptions. The primary audience of the paper includes pattern writers interested in pattern mining using LLMs.

The paper is organized as follows. First, I review existing approaches to pattern mining, human-AI collaboration with ChatGPT, and patterns for LLMs. I then describe an eight-step process for mining patterns from known uses with ChatGPT. Next, I present the application of the process to create a pattern language for integrating LLMs with data sources and tools. I then describe the patterns that were created and subsequently revised using ChatGPT. This is followed by a pattern story showing the application of the patterns to one of the examples. I conclude with a discussion of the approach, and directions for future research. Appendix~\ref{sec:prompts} contains the prompts used to mine the patterns. A log of the conversation with ChatGPT is available online (see Appendix~\ref{sec:log}).

\section{Literature Review}

This section reviews existing approaches to pattern mining, then discusses examples of human-AI collaboration with ChatGPT, and finally summarizes existing work on patterns for LLMs. Note that the goal of this section is to provide a quick overview of related work, not a thorough review of the literature.

\subsection{Pattern Mining}

Iba \& Isaku \citeN{iba2012holistic} describe a holistic pattern mining approach that includes element mining, visual clustering, and the discovery of pattern “seeds”. Hanmer \citeN{hanmer2012pattern} documents patterns that facilitate the pattern writing and review process. Iba \& Isaku \cite{iba2016pattern} present a detailed pattern language for mining Iba-style patterns. Weiss \citeN{weiss2017evidence} describes an empirical approach to discovering patterns from design documentation using word clouds to help identify pattern elements. Linden \& Cybulski \citeN{linden2009hermeneutics} show how the hermeneutic method can be used to induce patterns from interviews with designers. These existing approaches do not use AI.

\subsection{Human-AI Collaboration with ChatGPT}

De Paoli \citeN{depaoli2023inductive} describes an exploratory approach to conducting thematic analysis using LLMs. This paper is interesting for two reasons. It is an example of a hybrid approach that involves the collaboration between human researchers and an AI. Pattern mining and thematic analysis also involve similar inductive analysis steps. Schmidt \citeN{schmidt2023pattern} share their experience using ChatGPT to discuss and refine a pattern description.

\subsection{Patterns for LLMs}

White \citeN{white2023prompt} document a catalog of patterns for prompt engineering, which the authors see as a way to program an LLM through prompts. Schukart \citeN{schuckart2023interdisciplinary} describes patterns for the effective use of generative AI in a work environment. However, patterns for integrating LLMs with data and tools have yet not been documented. I felt it would be a suitable test case for my approach.

\section{Process for Mining Patterns with ChatGPT}

In the proposed process, pattern mining involves a collaboration between domain expert and ChatGPT, in which human skills are augmented by those of the AI. Identifying the steps of the process and the particular prompts to be used at each stage required some experimentation. Table~\ref{tab:steps-process} shows the steps of the process and their objectives. These steps were conceptualized by the author and iteratively revised during the experiment. However, we used ChatGPT to help produce the initial descriptions of the steps by asking it to summarize the process from the conversation log.% 
\footnote{This is a useful capability of ChatGPT. It can ``reflect'' on its conversation with the user, e.g., by breaking down the conversation into its key steps. To create a summary of the steps of the process that emerged from the conversation, I used the following prompt: {\sf Outline the process we used to co-create these patterns.}}

\begin{table*}[p]
  \caption{Steps of the pattern mining process}
  \label{tab:steps-process}
  \begin{tabular}{p{1.5in}p{4.5in}}
    \toprule
    Step & Objective \\
    \midrule
    Identify initial examples & Establish a concrete basis for discussing how the components (or more precisely, their capabilities) are combined in real-world applications through a series of examples. \\
    Extract common solutions & Identify the recurring solutions used in different examples to solve common problems. \\
    Define problems & Clarify the specific problems or challenges that each solution addresses. This provides a basis for understanding the applicability and value of the solutions. \\
    Distill problem-solution pairs into patterns & Formalize the common solutions and their corresponding problems into recognizable and reusable patterns that can be applied in similar contexts. \\
    Identify affordances & Understand the intrinsic affordances of each component that can be harnessed in these patterns, providing a deeper insight into why certain patterns are effective. \\
    Relate patterns to affordances & Demonstrate how each pattern harnesses the inherent affordances of these components, providing a better understanding of the applicability and effectiveness of the pattern. \\
    Refine iteratively & Ensure that the patterns are described accurately and thoroughly. This includes exploring dependencies between the patterns and refining the descriptions. \\
    Consolidate the patterns & Provide a cogent documentation for understanding and applying these patterns in future system designs. This includes creating pattern stories for application examples. \\
    \bottomrule
  \end{tabular}
\end{table*}

The process begins with the selection and documentation of application scenarios to be used as the initial examples, or known uses, from which the patterns will be mined. These scenarios should be provided by a domain expert, and must contain sufficient detail about the design goal and the key steps of the application design.

There should be some overlap as well as some variation between these examples, so that the examples cover the space of design choices that we want the patterns to capture. If we conceptualize a solution as a combination of capabilities offered by the underlying components (in the case of an LLM, a component, an example of a capability would be its ability to synthesize content), then each example should show a slightly different combination of capabilities. 

The purpose of the next two steps (extracting common solutions, and defining problems) is to infer the solutions and their corresponding problems inductively in collaboration with ChatGPT. These problem-solution pairs can then be distilled into patterns. For each pattern, we extract the context and forces, and capture the known uses of the solution from the examples. The output of this step is short forms of the patterns. These can later be elaborated into more detailed versions of the patterns in the refinement step.

The goal of the next step is to extract the affordances (i.e., key functionalities or capabilities) of the application components. This will give us a deeper understanding of the components and how they interact in the examples and the patterns. In this context, “affordances” refers to the inherent features, capabilities, or functionalities of components that can be used in the design of a system. Affordances determine which capabilities can be selected when integrating these components into solutions or applications.

Following affordance actualization theory \cite{volkoff2017affordance}, a theory for how to design with affordances, each solution can be thought of as a configuration of affordances. For an analogy, think of cooking a meal. This is the equivalent of the solution. To prepare the meal, you need ingredients. Those are the components of the solution. Each ingredient has properties such as flavor or nutritional value. Those are the affordances. 

Relating these affordances to patterns allows us to understand which patterns use which affordances and how. We can make the affordances used by a pattern explicit by including them in its description. A good place to do this is in the body of the pattern’s solution. Affordances, thus, provide a common language for describing the solutions of patterns within a pattern language. Traditionally, a pattern describes only the structure of the solution, that is, its components and their relationships, not their affordances.

The final two steps involve refining and consolidating the patterns. Refining patterns comprises identifying dependencies between them, renaming patterns, or adding more information to pattern descriptions such as citations or solution details. Consolidating the patterns includes studying patterns in the context of solutions. This can be done by creating pattern stories \cite{buschmann2007pattern}. It also involves manually completing the pattern descriptions to add more detail beyond the scope of the examples.

Each of these steps can be supported by prompts. Examples of the use of these prompts are given in the next section. A complete list of the prompts can be found in Table~\ref{tab:prompts} in Appendix~\ref{sec:prompts}).

\section{Application of the Pattern Mining Process}

The following sections describe an application of the proposed pattern mining process. It is organized around the steps from Table~\ref{tab:steps-process}, and provides examples of the prompts used to implement the steps. Prompts are visually highlighted by putting them inside a box.

\subsection{Identify Initial Examples}

The examples of applications from which the patterns were mined are shown in Table~\ref{tab:examples}. For this experiment, the examples were applications implemented by the author as illustrative use cases of applying LLMs. The examples were chosen because they share certain elements, but also exemplify unique design challenges. Here we chose three examples, but more examples may be required.

\begin{table*}[p]
  \caption{Application scenarios used as input to the pattern mining process}
  \label{tab:examples}
  \begin{tabular}{p{1.5in}p{4.5in}}
    \toprule
    Application & Description \\
    \midrule
    Customer support scenario & The first example is a customer support scenario. In this scenario, the goal is to use an LLM to answer frequently asked questions (FAQs). The process starts by parsing a structured document with FAQs and splitting it into questions and answers. These answers are then indexed in a vector database using the associated questions as keys. When queried, the database returns answers that closely match the query based on semantic relevance. When a customer asks a question, the system retrieves the most semantically relevant answers from the vector database. The LLM then uses these answers as context to create a response. If the LLM determines that none of the answers addresses the customer's question, it will reply that there is no suitable answer. \\
    \midrule
    Research assistant & The second example is a system designed to act as a research assistant. It facilitates the creation of a literature review by integrating a search engine, a topic modeling tool, and an LLM. The process begins by retrieving abstracts of research articles that match user-supplied search terms from arXiv.org via its API. Next, the topic modeling tool is called on to categorize these abstracts into a predetermined number of topics. For each topic, a list of keywords and a set of abstracts that best represent the topic are returned. The LLM is then prompted to assign labels to these topics and generate descriptive summaries for them. Finally, the LLM is asked to synthesize the content from the most representative abstracts for each topic, crafting them into a cohesive narrative for each stream of the review. \\
    \midrule
    Information extraction & In the third example, the focus is on using LLMs for information extraction. As an example, consider the task of analyzing stories of startup failures to understand why startups fail. The process starts with scraping stories of startup failures from a CB Insights research report on startup failure post-mortems. The LLM is then asked to extract the reasons behind the failures from each story. These extracted failure reasons are converted into embeddings using the LLM. A clustering algorithm (for example, k-means) is applied to these embeddings to group similar failure reasons together into clusters. Finally, the LLM synthesizes the failure reasons in each cluster into a description of the cluster. \\
    \bottomrule
  \end{tabular}
\end{table*}

To extract the common solutions, I provided the examples and then instructed ChatGPT to identify the recurring solutions common to those examples. The prompt begins with:

\begin{shaded*}
\noindent In the following, you will help me to mine patterns for the integration of Large Language Models (LLMs), data sources, and external tools through a structured, iterative dialog. I will start by providing examples of how LLMs, data sources, and external tools can be integrated in real-world applications. In each case, the integration is achieved by an application that defines the overall logic of how the LLM, data sources, and external tools are combined and calls on them as needed.
\end{shaded*}

This is followed by the text of the application scenarios from Table 2. The prompt ends with:

\begin{shaded*}\noindent What are the recurring solutions across these examples?\end{shaded*}

The result is a list of solutions and their descriptions (for brevity, just the names of the solutions are shown below):

\begin{shaded*}\noindent Data Preprocessing\\
Data Structuring and Enhancement\\
Integration with External Tools\\
Semantic Understanding and Synthesis\\
Iterative Refinement and Feedback\\
Custom Application Logic\\
Adaptive Response Generation\end{shaded*}

Here is the description of one of the solutions:

\begin{shaded*}\noindent Custom Application Logic: Each example is driven by custom application logic that defines how the LLM, data sources, and external tools are integrated. This logic includes the sequence of operations, the criteria for data retrieval and processing, and the way in which the LLM's outputs are used. The application logic is tailored to the specific goal of the system, whether it's providing customer support, assisting with research, or analyzing startup failures.\end{shaded*}

\subsection{Define Problems}

With the recurring solutions identified, I then asked ChatGPT to identify the common problems underlying the solutions:

\begin{shaded*}\noindent For each solution, what is the common problem underlying this solution?\end{shaded*}

The problem corresponding to the solution above was:

\begin{shaded*}\noindent Custom Application Logic: This solution addresses the problem of the diverse and specific requirements of different applications. A one-size-fits-all approach is often insufficient for the wide range of tasks and goals that LLMs can be applied to, necessitating custom logic that tailors the integration of data, tools, and LLM capabilities to the specific needs of each application.\end{shaded*}

\subsection{Distill Problem-Solution Pairs into Patterns}

Next, I asked ChatGPT to compile the problem-solution pairs and describe them as patterns. The following prompt defines the expected output format by specifying the sections of a pattern:

\begin{shaded*}\noindent For each problem-solution pair, describe the solution in the following format:\\
Context: in which situation does the problem occur?\\
Problem: succinct statement of the problem\\
Forces: elaboration on what makes the problem challenging\\
Solution: succinct statement of the solution\\
Known uses: how do the examples illustrate the solution?\end{shaded*}

Note that this template does not include a resulting context section. This section will be added during the refinement step. The resulting context should also include the related patterns that help address the consequences of the pattern. However, we cannot identify the relationships between the patterns until all the patterns have been described. These pattern descriptions created in this step can then be provided to ChatGPT as context.

Here is the output generated from the problem and solution for the Custom Application Logic pattern:

\begin{shaded*}\noindent Custom Application Logic\\
Context: Diverse applications of LLMs, each with unique goals, data sources, and requirements, from customer support to research assistance to failure analysis.\\
Problem: A one-size-fits-all approach does not adequately address the specific requirements of different applications.\\
Forces: The diversity of application domains, the specificity of user needs, and the variability of data sources contribute to the complexity.\\
Solution: Develop custom logic that defines the specific integration and interaction of LLMs, data, and tools tailored to the application's goals.\\
Known Uses: The specific sequences of data retrieval, processing, and LLM interaction in each example illustrate custom application logic tailored to different goals.\end{shaded*}

\subsection{Identify Affordances}

In this step, I used the following prompt to help me identify the affordances of the components. To guide ChatGPT and as a way of defining what we mean by affordances, the goal of this step is explicitly stated as part of the prompt.

\begin{shaded*}\noindent Next, I want you to help me identify affordances. The goal is to gain a deeper understanding of the patterns and how they work. What affordances of LLMs, databases and external tools are actualized in these patterns?\end{shaded*}

The response to this prompt is a list of the affordances for each component. The result is better visualized as a table, see Table~\ref{tab:affordances}. These are affordances that can be actualized or realized by solutions, as well as, applications that use those solutions.

\begin{table*}[p]
  \caption{List of affordances for each component}
  \label{tab:affordances}
  \begin{tabular}{lp{2in}p{3.5in}}
    \toprule
    Component & Affordance & Description \\
    \midrule
    LLM & Natural language understanding & LLMs can understand and interpret human language \\
     & Content generation & LLMs can generate coherent, contextually appropriate text \\
     & Semantic search and matching & LLMs can go beyond keyword matching to find relevant content based on the meaning and context \\
     & Adaptive learning & LLMs can adapt responses based on new information \\
    \midrule
    Database & Structured data storage & Databases afford the structured storage of large volumes of data, enabling efficient retrieval and management of information \\
     & Semantic indexing and retrieval & Advanced databases, such as vector databases, support the storage and retrieval of information based on semantic similarity \\
     & Scalability and performance & Databases are designed to handle scaling \\
     & Data organization and categorization & Databases can organize data in a variety of ways (e.g., relational, graph-based, etc.) to support complex data relationships \\
    \midrule
    External tool & Specialized analytical capabilities & External tools offer analytical capabilities that LLMs lack \\
    & Data preprocessing and enhancement & Data scraping, cleaning, and preprocessing tools enable the transformation of raw data into formats suitable for LLMs \\
    & Efficiency and optimization & Many external tools are optimized for specific tasks (e.g., search) \\
    & Interoperability and integration & External tools often come with APIs or integration capabilities \\
    \bottomrule
  \end{tabular}
\end{table*}

\subsection{Relate Patterns to Affordances}

Now, we are ready to determine which patterns use or build on which affordances by asking:

\begin{shaded*}\noindent Cross-reference the affordances with the patterns by identifying which affordances are used by which patterns.\end{shaded*}

This output lists the affordances actualized by each pattern. To make it easier to view the output, these affordance can then be collated into a table (Table~\ref{tab:cross-references}). Each entry of the table represents an affordance and which patterns it is used in. For example, for the Custom Application Logic pattern the output is:

\begin{shaded*}\noindent LLM Affordances:\\
Content Generation: LLMs can be programmed with custom logic to generate specific outputs required by the application.\\
Database Affordances:\\
Data Organization and Categorization: Databases can be structured according to the custom logic of the application to optimize data retrieval and storage.\\
External Tool Affordances:\\
Interoperability and Integration: Essential for custom applications that integrate multiple tools and systems, enabling seamless workflows.\end{shaded*}

Table~\ref{tab:cross-references} shows the affordances cross-referenced with the patterns that use them. Based on the information in this table, we will add affordances to the solution section of the pattern descriptions. Documenting the affordances actualized by each pattern allows us to describe a solution in terms of the capabilities of the underlying components it uses or builds on.

\begin{table*}[p]
  \caption{Affordances cross-referenced with the patterns that use them}
  \label{tab:cross-references}
  \begin{tabular}{llccccccc}
    \toprule
    Component & Affordance & \multicolumn{7}{c}{Pattern} \\
    \midrule
    & & \rotatebox{90}{\parbox{1.25in}{Data Retrieval and \\ Preprocessing}} 
      & \rotatebox{90}{\parbox{1.25in}{Data Structuring and \\ Enhancement}}
      & \rotatebox{90}{\parbox{1.25in}{Integration with \\ External Tools} }
      & \rotatebox{90}{\parbox{1.25in}{Semantic Understanding \\ and Synthesis}}
      & \rotatebox{90}{\parbox{1.25in}{Iterative Refinement \\ and Feedback} }
      & \rotatebox{90}{\parbox{1.25in}{Custom Application \\ Logic}}
      & \rotatebox{90}{\parbox{1.25in}{Adaptive Response \\ Generation}} \\
    LLM 
      & Natural language understanding &  &  &  & X &  &  & X \\
      & Content Generation & X &  &  &   &  & X & X \\
      & Semantic Search and Matching &  & X &  &  &  &  & \\
      & Adaptive Learning &  &  & X &  & X &  & \\
    Database 
      & Structured Data Storage &  & X &  &  &  &  &\\
      & Semantic Indexing and Retrieval &  & X &  &  &  &  &\\
      & Scalability and Performance &  &  &  &  & X &  &\\
      & Data Organization and Categorization &  &  &  &  &  & X &\\
    External Tool 
      & Specialized Analytical Capabilities &  & X & X &  &  &  &\\
      & Data Preprocessing and Enhancement & X &  &  &  &  &  &\\
      & Efficiency and Optimization &  &  &  &  &  &  &\\
      & Interoperability and Integration &  &  &  &  &  & X & \\
    \bottomrule
\end{tabular}
\end{table*}

\subsection{Refine Iteratively}

Dependencies between the patterns can be explored with the help of the prompt below. This prompt includes instructions for what we mean by pattern dependencies and how to identify them.

\begin{shaded*}\noindent There are some possible dependencies between the patterns, where you may want to use another pattern after applying a pattern. For the patterns in the list above, identify the patterns that should be considered next after applying the pattern, if any. Also include a reason why this pattern should be considered in the context of this pattern. The result should be captured in the Resulting Context section of the pattern. Don't repeat the whole patterns, but just give me the Resulting Context sections.\end{shaded*}

Dependencies will be documented in the resulting context section of a pattern. For example, the generated resulting context for the Custom Application Logic pattern is:

\begin{shaded*}\noindent Resulting Context: After establishing custom application logic, Integration with External Tools may be revisited to ensure that all necessary tools are effectively integrated into the workflow. This ensures that the application logic fully leverages the capabilities of each component, from data handling to analysis and response generation.\end{shaded*}

\section{Pattern Language}

This section describes a pattern language for integrating LLMs with data sources and tools, which was created using the process described in the previous section. The motivation for these patterns is that LLMs need to be integrated with external data sources and tools to realize their full potential. Those data sources and tools augment the capabilities of an LLM. Table~\ref{tab:pattern-language} summarizes the patterns in this pattern language.

\begin{table*}[p]
  \caption{Summary of the patterns}
  \label{tab:pattern-language}
  \begin{tabular}{lp{4in}}
    \toprule
    Patterm & Description \\
    \midrule
    Data Preprocessing & 
    Implement preprocessing steps to clean, normalize, and structure raw data into a usable format. \\
    Data Structuring and Enhancement &
    Structure and enhance data using techniques like semantic indexing, embedding, or categorization to add meaningful organization and context. \\
    Tool Integration &
    Augment LLMs by integrating external tools that provide specialized functions or capabilities not inherent to LLMs. \\
    Semantic Understanding and Synthesis &
    Harness LLMs for their deep semantic understanding and synthesis capabilities to generate coherent and contextually relevant content. \\
    Adaptive Response &
    Design LLMs to generate adaptive responses, either by acknowledging limitations or by providing the best possible alternative response. \\
    Custom Logic &
    Develop custom logic that defines the specific integration and interaction of LLMs, data, and tools tailored to the goals of the application. \\
    \bottomrule
\end{tabular}
\end{table*}

The patterns are described using the Alexandrian format \cite{alexander1977pattern}. It begins with the name of the pattern, and is followed by its context. The problem statement (in bold) is separated from the context by stars. The problem statement is elaborated by the forces. The word “Therefore” separates the forces from the solution. The solution consists of a solution statement (in bold) and (optional) solution details. Separated from the solution by stars is the resulting context. It describes the consequences of applying the solution and (optionally) any related patterns.

The affordances actualized by the solution could be described in different sections of the pattern. One option, adopted here, is to describe the affordances and how they are used by the pattern in the detailed solution section. Taken together, they provide an explanation of the solution. Visually, the affordances are highlighted by putting them in italics. Another option would be to present the affordances as a list of keywords, such as in the margin next to the pattern description for easy reference.

\subsection{Data Preprocessing}

Data sources provide data in a variety of formats.

\stars

{\bfseries Raw data is often unstructured, noisy, and not in a directly usable format.}

The diversity of data sources and formats, the presence of irrelevant or redundant information, and the need for data to be in a specific structure for analysis or processing make this problem challenging.

Therefore,

{\bfseries Implement preprocessing steps to clean, normalize, and structure raw data into a usable format.}

{\itshape Data preprocessing and enhancement}, including scraping, cleaning, and formatting raw data, is crucial to making the data usable for analysis by LLMs or storage in databases. The {\itshape content generation} capabilities of an LLM can also be used to preprocess text and normalize or summarize information before it is indexed in a database or further analyzed in an LLM.

\stars

In the customer support example, FAQs are parsed and structured for indexing. In the research assistant scenario, research abstracts are retrieved and preprocessed for topic modeling.

After data is retrieved and preprocessed, consider applying {\itshape Data Structuring and Enhancement} to organize and enhance the data to make it more suitable for analysis or querying. This is critical because structured, enhanced data can be more effectively indexed, searched, and analyzed, especially when dealing with complex queries or the need for semantic understanding.

\subsection{Data Structuring and Enhancement}

Raw or preprocessed data lacks organization, categorization, or semantic richness.

\stars

{\bfseries Data is unstructured and lacks semantic organization, making it difficult to analyze or query effectively.}

The inherent complexity of natural language, the variety of possible interpretations, and the need for semantic understanding to perform specific tasks contribute to the challenge.

Therefore,

{\bfseries Structure and enhance data using techniques like semantic indexing, embedding, or categorization to add meaningful organization and context.}

To organize preprocessed data in an accessible manner, it is essential to structure the data ({\itshape structured data storage}), for example, as a list or in JSON format. {\itshape Semantic indexing and retrieval} can be used to enhance the retrieval process, making it more relevant and efficient. LLMs can help structure data by identifying semantic relationships or categorizations, which can, in turn, inform database indexing ({\itshape semantic search and matching}). The {\itshape specialized analytical capabilities} of tools such as topic modeling algorithms can also be used to structure and categorize text based on its content.

\stars

The vector database indexing of FAQs in the customer support example and the categorization of research abstracts into topics in the research assistant example demonstrate this solution.

Once data is structured and enhanced, {\itshape Semantic Understanding and Synthesis} often follow, where LLMs can use the organized data to generate insights, summaries, or responses.

\subsection{Tool Integration}

LLM capabilities need to be augmented with specialized functions for analysis or processing.

\stars

{\bfseries Despite their versatility, LLMs cannot handle all specialized tasks required by specific applications.}

The specialized nature of certain tasks, the limitations of LLMs in specific analytical domains, and the need for processing efficiency and scalability in make this a challenging problem.

Therefore,

{\bfseries Augment LLMs by integrating external tools that provide specialized functions or capabilities not inherent to LLMs.}

LLMs can adjust their responses based on output from external tools ({\itshape adaptive learning}). The tools provide {\itshape specialized analytical capabilities} that LLMs do not inherently have.

\stars

The use of a topic modeling tool in the research assistant scenario and a clustering algorithm in the startup failure analysis example illustrate this integration.

The output of the tool may not be directly suitable for analysis, requiring {\itshape Data Processing} to transform the output into a suitable format that can be consumed by further processing steps.

\subsection{Semantic Understanding and Synthesis}

An application requires deep understanding of context and generation of coherent, contextually appropriate content.

\stars

{\bfseries Simple models fail to capture the nuances of language and context, leading to superficial or irrelevant responses.}

The complexity of natural language, the subtlety of context, and the need for nuanced understanding and synthesis in responses add to the challenge.

Therefore,

{\bfseries Harness LLMs for their deep semantic understanding and synthesis capabilities to generate coherent and contextually relevant content.}

{\itshape Natural language understanding} and {\itshape content generation} are at the heart of this pattern, enabling LLMs to understand context and generate coherent, contextually relevant content.

\stars

The generation of topic labels and summaries in the research assistant example and the synthesis of cluster descriptions in the analysis of startup failures demonstrate this capability.

Following semantic understanding and synthesis, consider {\itshape Adaptive Response} to ensure that the system can handle edge cases or queries outside its training range. This is important to maintain user engagement and trust by providing relevant responses even when definitive answers are not available.

\subsection{Adaptive Response}

LLMs may encounter user queries that fall outside their training or the scope of available data.

\stars

{\bfseries Inevitable encounters with edge cases or limitations in data or LLM capabilities can lead to inadequate responses.}

The unpredictability of user queries, the inherent limitations of LLMs, and the finite scope of available data all contribute to this challenge.

Therefore,

{\bfseries Design LLMs to generate adaptive responses, either by acknowledging limitations or by providing the best possible alternative response.}

{\itshape Natural language understanding} enables an LLM to recognize when a user query falls outside its training or the scope of the data provided to it as part of the context. LLMs can craft responses that acknowledge limitations or provide the best possible alternative response ({\itshape content generation}).

\stars

The customer support example, where the LLM generates a response when no suitable FAQ answer is found, demonstrates adaptive response generation.

After adaptive response generation, revisit {\itshape Semantic Understanding and Synthesis} to further refine the system's understanding and output based on user interactions and feedback. This could be achieved, for example, by fine-tuning the LLM or adding in-context examples. This cycle improves the system's ability to generate even more contextually appropriate and helpful responses over time.

\subsection{Custom Logic}

Applications of LLMs cover a wide range of domains, user needs, and data sources.

\stars

{\bfseries A one-size-fits-all approach does not adequately address the specific requirements of different applications.}

The diversity of application domains, the specificity of user needs, and the variability of data sources contribute to the application’s complexity.

Therefore, 

{\bfseries Develop custom logic that defines the specific integration and interaction of LLMs, data, and tools tailored to the goals of the application.}

LLMs can be prompted to generate application-specific outputs required by the application ({\itshape content generation}). Data sources can be structured according to the custom logic of the application ({\itshape data organization and categorization}). Finally, {\itshape interoperability and integration} (for example, through APIs) are essential to enable the integration of multiple external tools and systems.

\stars

The specific sequences of data retrieval, processing, and LLM interaction in each example illustrate how custom application logic is tailored to meet different objectives.

After the custom application logic is created, revisit {\itshape Tool Integration} to ensure that all the necessary tools are effectively integrated into the workflow. Specifically, focus on the following aspects: data transformation (ensuring that each tool receives information in the required format), task alignment (matching the most appropriate tool to each task), and task granularity (the idea that each tool should perform a specific task or a small set of related tasks to improve maintainability).

\section{Pattern Story}

The following pattern story shows the patterns exemplified by the Research Assistant application scenario. The draft of the story was generated by using the following prompt:

\begin{shaded*}\noindent Here is a recap of the Research Assistant example. Please identify the relevant patterns.\\
This example is a system designed to act as a research assistant. It facilitates the creation of a literature review by integrating a search engine, a topic modeling tool, and an LLM. The process begins by retrieving abstracts of research articles that match user-supplied search terms from arXiv.org via its API. Next, the topic modeling tool is called on to categorize these abstracts into a predetermined number of topics. For each topic, a list of keywords and a set of abstracts that best represent the topic are returned. The LLM is then prompted to assign labels to these topics and generate descriptive summaries for them. Finally, the LLM is asked to synthesize the content from the most representative abstracts for each topic, crafting them into a cohesive narrative for each stream of the review.\end{shaded*}
 
This example integrates several patterns in the use of LLMs, databases, and external tools to facilitate the creation of a literature review. Here are the relevant patterns identified for this example:

\begin{enumerate}
\item Data Preprocessing:
The system retrieves article abstracts from arXiv.org using its API. The abstracts need to be extracted from the search results returned by the API.
\item Data Structuring and Enhancement: 
The topic modeling tool groups the retrieved abstracts into topics, structuring the data into a more organized form that enhances its usability for the task at hand. This involves semantic understanding and categorization of the content of the abstracts.
\item Tool Integration:
The system integrates a search engine (ar\-Xiv.org's API) for data retrieval and a topic modeling tool for data categorization. These tools augment the capabilities of the LLM by providing specialized functionality not inherent to the LLM, such as accessing a database of research articles and categorizing abstracts into coherent topics.
\item Semantic Understanding and Synthesis:
The LLM is used to assign labels to topics, demonstrating its capability for semantic understanding. In addition, the LLM synthesizes the content from the most representative abstracts for each topic into cohesive narratives, demonstrating its content generation and synthesis capabilities.
\item Custom Logic:
The entire process, from data retrieval to the generation of a literature review narrative, is driven by custom logic that defines how the LLM, data sources, and tools are combined. This logic includes the sequence of steps from database query to content synthesis.
\end{enumerate}

\section{Discussion}

In this exploratory paper, I proposed an approach to pattern mining that involves a collaboration between a human domain expert and an AI (ChatGPT). In this section, I discuss how well ChatGPT was able to extract patterns from known uses, as well as the necessary changes to the initial pattern descriptions generated by ChatGPT that would be required by a domain expert.

\subsection{Ability to Extract Patterns from Examples}

The quality of the examples and the variety of design choices they represent will have a significant impact on the patterns that can be extracted. Also, to a large extent, patterns will only be generated for design choices that are reflected in the examples. Domain experts will be aware of other patterns and design choices, but these may not be directly supported by evidence from the examples.

ChatGPT is also known to occasionally produce generic answers, especially when asked specific questions that go beyond the data on which it was trained. This can also be observed in the patterns it generates. Out of seven patterns in this experiment, ChatGPT suggested one pattern that was quite general (Iterative Refinement and Feedback). While, iteration is obviously a widely used principle, including such a pattern in the pattern language does not strengthen the language. This pattern was also not supported by the pattern stories generated for the examples.

\subsection{Necessary Changes to the Pattern Descriptions}

The output produced by ChatGPT should be taken as a starting point for the pattern descriptions. Pattern mining with ChatGPT should not be seen a one-way street where we simply accept the output it produces, but as a co-creation process where the AI augments our human ability.\footnote{In a recent book, Mollick \citeN{mollick2024cointelligence} refers to this human-AI collaboration process as co-intelligence.} Thus, instead of asking ChatGPT to combine the various pattern elements it produced (initial description, resulting context, and affordances) into a complete pattern description, I chose to combine them manually.\footnote{I initially experimented with asking ChatGPT to expand the pattern descriptions given all the additional information. However, the result was wordy and formulaic.}
 
This gave me a chance to digest what ChatGPT had produced, and "humanize" the pattern descriptions, which meant reflecting them through the lens of a domain expert, correcting errors and inconsistencies, and adding explanations and my own interpretations. Besides general editorial changes to improve readability, there were also recurring specific changes, as discussed below.

Pattern names generated by ChatGPT were often too long to be useful as handles for referring to the patterns. For example, Custom Application Logic was renamed to Custom Logic.

The context element required frequent tweaking. ChatGPT’s output was structured in the format “when some need arises", citing specific examples instead of a general description of the setting. For instance, the suggested context for Data Preprocessing read: “When raw data is needed for analysis or processing in applications like customer support systems, research assistants, or startup failure analysis.” This was generalized by making it explicit that the data could come in a variety of formats.

Occasionally, I had to add a missing resulting context. For instance, there was no resulting context for the Tool Integration pattern. However, since tools produce data in formats that may not be suitable for processing by the LLM, this data must first be processed by applying the Data Processing pattern. The explanation given in the resulting context was also sometimes lacking in detail. For example, in Adaptive Response, I added this explanation for why Semantic Understanding and Synthesis should be considered next: “This could be achieved, for example, by fine-tuning the LLM or adding in-context examples.” This is a necessary detail that was not highlighted as part of the known uses.
 
Sometimes, the terminology suggested by ChatGPT was also imprecise. For example, ChatGPT wrote that LLMs can be "programmed", when the term "prompted" would be more accurate. This is to be expected, as ChatGPT does not always have the necessary domain knowledge. As a statistical model, it might also suggest a word that is commonly used, but not correct in a specific context.

Furthermore, some suggestions were too specific. For example, ChatGPT provided this explanation of an affordance: "Tools like topic modeling algorithms can structure and categorize data based on content, which can then be indexed in databases." However, the power of topic modeling is not limited to its use for indexing. While a topic model can be used to assign documents to topics, the output (e.g., the keywords associated with a topic) can also be analyzed by an LLM or another tool.

I did not use ChatGPT to add to the pattern descriptions, such as expanding the discussion of the forces or adding implementation details to the solution. Keeping the generated pattern descriptions concise gives human domain experts more opportunity to flesh out the patterns with their own insights into the domain. However, it might be interesting to explore how to design prompts to enhance the information added by domain experts, for example, by identifying gaps in their thinking.

\section{Conclusion}

This paper describes the results of an experiment using ChatGPT (GPT 4) for pattern mining. The initial results are promising, however, in its current form the proposed process has limitations that should be addressed by future work.    

\subsection{Limitations}

There are three potential limitations of the current process: in how broadly the process is applicable, in how known uses are made available to the LLM for processing, and in the how the outcomes depend on the quality of the prompts. I will discuss each in turn.

\subsubsection*{Applicability of the process}

Since this paper describes only a single case study of the use of this process, it is difficult to generalize how widely applicable it is. Instead, I will offer tentative answers to the following questions: Is the process better suited to certain types of patterns or domains than for others? What are the characteristics of a domain that suggest that the process can or cannot be applied? 

My sense is that the process is more applicable to contexts where the known uses are described in a narrative format, rather than in (raw) code or design diagrams. Such a narrative format would be expected to include a rationale for the various design decisions illustrated by the application scenarios. Code or design diagrams, on the other hand, often do not include an explanation of why the solutions they embody work, although, they may, of course do so (in particular, if the code is properly commented).

Thus, the process should be more appropriate for domains that involve architectural design decisions, processes (e.g., in software development), or organizational design choices. Initial experiments with an earlier version of the process, where I tried to rediscover patterns for agile practices from a description of their known uses, suggest that the process is also applicable to such contexts. However, more work (as discussed in the Future Work section) is needed to draw more solid conclusions.

\subsubsection*{Making known uses available}

Currently, the full description of the known uses is included in the context of the prompt. However, LLMs have limits on how much context can be provided as part of a prompt. Technically, the number of words or tokens that can be used to describe the context is limited. While larger contexts can be accommodated as LLMs evolve, there are also known issues with how much attention an LLM pays to information in the context. This can lead to an uneven processing of the information. Thus, not all known uses may be given the same weight during processing.

A solution to this issue could come from the recent progress in conducting thematic analysis using LLMs \cite{depaoli2023inductive}. In this approach, known uses could be annotated with codes (e.g., marking a sentence as related to a problem or a solution), and the coded sections could then be fed into the LLM for further processing instead of the full, unprocessed known uses. Such an approach would allow the proposed process to scale to larger numbers of known uses.

\subsubsection*{Quality of the prompts}

The prompts identified may not be optimal or as reliable as alternative prompts. The prompts could also incorporate more explicit knowledge about patterns and writing patterns, rather than relying on the definitions available to ChatGPT.

\subsection{Future Work}

As an exploratory paper, this paper does not pretend to have all the answers. There is much room for improvement and expansion of the work described. Some of these opportunities are identified next in the form of future experiments to be conducted.

One interesting experiment would be to compare the granularity of the patterns extracted with the patterns extracted by domain experts when given the same known uses. For example, in the case study (integrating LLMs with data and tools), one might have expected two patterns (Tool Integration and Data Integration), instead of one pattern (Tool Integration) that covers both cases, treating data sources as a type of tool. However, one could also argue that this choice has some validity and that the two patterns would necessarily have significant overlap.

A variant of this experiment would be to apply the process to an existing pattern language, starting with the known uses and examples documented in the pattern language. In this way, one could compare the output produced by the process with the actual patterns in the pattern language. Such an experiment would focus on the completeness of the pattern language rather than the granularity of the patterns.

A related experiment would be to repeat the prompts several times, starting from the same examples, and observe what differences, if any, are produced, and whether the differences are significant. A difference would be considered significant if one experiment produces a pattern that a domain expert would consider essential, and another experiment does not produce a similar pattern. A difference would not be considered significant, if the patterns merely "partition" the space of design choices in different ways. Such differences are, at least to some extent, a matter of preference.

We could also build on existing work using thematic analysis for pattern mining such as the hermeneutic approach by \cite{linden2009hermeneutics}, and combine it with the recent work on conducting thematic analysis using ChatGPT \cite{depaoli2023inductive}. As noted above, this would make the proposed process more scalable by allowing it to include a larger and richer set of known uses.

Another aspect to be discussed in future work is the effectiveness of using affordances to describe the details of a solution. Can the concept be generally applied or is limited to domains where we can model pattern solutions in terms of components.

A low-hanging fruit is also to experiment with modifications to the prompts themselves. For instance, we might integrate explicit definitions of key concepts (e.g., patterns or affordances), or more specific instructions for conducting the steps of the process.

\begin{acks}
I would like to thank my shepherd Ademar Aguiar for his suggestions to improve this paper, especially by pointing out that the applicability of the process needs to be discussed. Thanks also go to the members of my workshop for their insightful discussion and feedback: Adrian Schuckart, Francisca Almeida, Luciane Adolfo, Tim Wellhausen, and Tomoki Kaneko. I learned a lot from you.
\end{acks}

\bibliographystyle{ACM-Reference-Format}
\bibliography{references}

\appendix

\section{Prompts}\label{sec:prompts}

The full prompts used during the collaboration with ChatGPT are listed in Table~\ref{tab:prompts}. They are grouped by process stage.

\begin{table*}[p]
  \caption{Summary of the prompts}
  \label{tab:prompts}
  \begin{tabular}{p{1.5in}p{4.5in}}
    \toprule
    Step & Objective \\
    \midrule
    Identify initial examples & {\sf In the following, you will help me to mine patterns for the integration of Large Language Models (LLMs), data sources, and external tools through a structured, iterative dialog. I will start by providing examples of how LLMs, data sources, and external tools can be integrated in real-world applications. In each case, the integration is achieved by an application that defines the overall logic of how the LLM, data sources, and external tools are combined and calls on them as needed.} \\ \\
    Define problems & {\sf What are the recurring solutions across these examples?} \\ \\
    Distill problem-solution pairs into patterns & \parbox{4.5in}{\sf For each solution, what is the common problem underlying this solution? \\ 
Context: in which situation does the problem occur? \\ 
Problem: succinct statement of the problem \\ 
Forces: elaboration on what makes the problem challenging \\ 
Solution: succinct statement of the solution \\ 
Known uses: how do the examples illustrate the solution?} \\ \\
    Identify affordances & {\sf Next, I want you to help me identify affordances. The goal is to gain a deeper understanding of the patterns and how they work. What affordances of LLMs, databases and external tools are actualized in these patterns?} \\ \\
    Relate patterns to affordances & {\sf Cross-reference the affordances with the patterns by identifying which affordances are used by which patterns.} \\ \\
    Refine iteratively & \parbox{4.5in}{\sf List the patterns identified so far.\\
Based on the examples, are there any patterns missing from the list.\\
There are some possible dependencies between the patterns, where you may want to use another pattern after applying a pattern. For the patterns in the list above, identify the patterns that should be considered next after applying the pattern, if any. Also include a reason why this pattern should be considered in the context of this pattern. The result should be captured in the Resulting Context section of the pattern. Don't repeat the whole patterns, but just give me the Resulting Context sections.} \\ \\
    Consolidate the patterns & \parbox{4.5in}{\sf Here is a recap of the X example. Please identify the relevant patterns.\\
Expand the short version of the Y pattern below.} \\
    \bottomrule
  \end{tabular}
\end{table*}

\section{Conversation Log}
\label{sec:log} 

The full log of the conversation with ChatGPT is available at: 

\noindent
\url{https://chatgpt.com/share/945f2f3a-f4d0-4f23-84b2-3939afe5ae7e}

\end{document}